\documentclass[letterpaper, 10 pt, journal, twoside]{IEEEtran} 

\IEEEoverridecommandlockouts                              

\usepackage{url}
\usepackage{float}
\usepackage{amssymb}
\usepackage{graphicx}
\usepackage{xcolor}
\usepackage{amsmath}
\usepackage{algorithm}
\usepackage{algpseudocode}
\usepackage{wrapfig}
\usepackage{booktabs}
\usepackage{caption}
\usepackage{subcaption}
\usepackage{cuted}

\usepackage[nosort,nocompress]{cite}

\usepackage[hidelinks,hyperfootnotes=false]{hyperref} 

\newcommand{\method}{EMMA}
\newcommand{\baseline}{Mobile ALOHA}

\title{EMMA: Scaling Mobile Manipulation via Egocentric Human Data}

\author{%
  Lawrence Y. Zhu$^{1}$, 
  Pranav Kuppili$^{1*}$, 
  Ryan Punamiya$^{1*}$, 
  Patcharapong Aphiwetsa$^{1\dagger}$, 
  Dhruv Patel$^{1\dagger}$,\\
  Simar Kareer$^{1}$, 
  Sehoon Ha$^{1}$, 
  Danfei Xu$^{1}$

\thanks{Manuscript received: August, 3, 2025; Revised November,
3, 2025; Accepted December, 12, 2025.}

\thanks{This paper was recommended for publication by Editor
Aleksandra Faust upon evaluation of the Associate Editor and
Reviewers' comments.}
  
\thanks{$^{1}$School of Interactive Computing, Georgia Institute of Technology, Atlanta, GA 30332, \{lawrencezhu, pkuppili3, rpunamiya6, paphiwetsa3, dpatel756, skareer6, sehoonha, danfei\}@gatech.edu}

\thanks{Digital Object Identifier (DOI): see top of this page.}
\thanks{$*,\dagger$ denote equal contributions}}

\begin{document}

\markboth{IEEE Robotics and Automation Letters. Preprint Version. Accepted Dec, 2025}
{Zhu \MakeLowercase{\textit{et al.}}: EMMA: Scaling Mobile Manipulation via Egocentric Human Data} 
\IEEEaftertitletext{\vspace{-7\baselineskip}}
\maketitle

\begin{strip}
\centering
\includegraphics[width=0.97\textwidth]{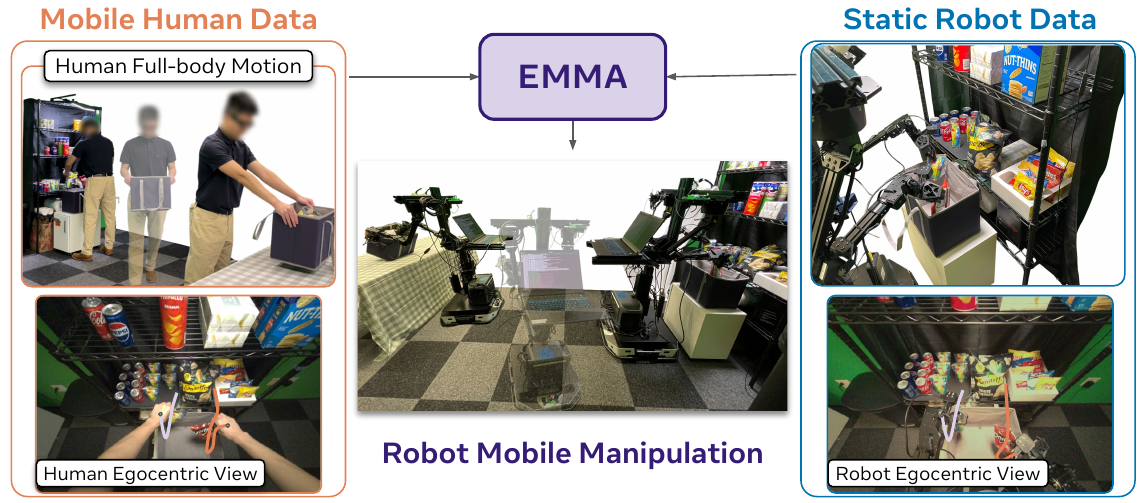}
\captionof{figure}{\method{} learns mobile manipulation policies \textit{without} collecting mobile manipulation teleoperation data. We achieve this through bridging embodiment kinematic gaps and unified co-training of mobile human data and static robot data.}
\label{fig:teaser_v2}
\vspace{-1em}
\end{strip}

\begin{abstract}

Scaling mobile manipulation imitation learning is bottlenecked by expensive mobile robot teleoperation. We present Egocentric Mobile MAnipulation (\method{}), an end-to-end framework training mobile manipulation policies from human mobile manipulation data with static robot data, sidestepping mobile teleoperation. To accomplish this, we co-train human full-body motion data with static robot data. In our experiments across four real-world tasks, \method{} demonstrates comparable performance to baselines trained on teleoperated mobile robot data (\baseline{}), achieving higher or equivalent task performance in full task success. We find that \method{} is able to generalize to new spatial configurations and scenes, and we observe positive performance scaling as we increase the hours of human data, opening new avenues for scalable robotic learning in real-world environments. Details of this project can be found at: \url{https://ego-moma.github.io}

\end{abstract}



\section{Introduction}

\IEEEPARstart{M}{obile} manipulation has emerged as one of the most challenging problems in robotics due to the dual demands of navigation and manipulation. While recent advances in robot policy learning have demonstrated impressive capabilities in static manipulation, extending these successes to mobile scenarios introduces substantial challenges. The primary obstacle is data scarcity; current approaches tackling mobile manipulation rely on teleoperation frameworks akin to \baseline{}~\cite{fu2024mobile} which face scalability limitations. This fundamental bottleneck limits dataset diversity and deployment robustness, particularly in unpredictable real-world settings where robots must navigate novel spatial configurations while maintaining manipulation precision.

Concurrently, recent work in cross-embodiment learning has explored human data as a scalable source of data for training end-to-end robot policies.  Critically, human video data is cheap to collect, does not require a physical robot, and can be collected ergonomically via XR wearables~\cite{egomimic, qiu2025humanoidpolicyhuman}. Human video data has been leveraged to train better visual representations~\cite{nair2022r3muniversalvisualrepresentation}, extract high-level object affordance~\cite{wang2023mimicplay, xu2024flowcrossdomainmanipulationinterface, wen2024anypointtrajectorymodelingpolicy, bharadhwaj2024track2actpredictingpointtracks}, or even as direct action supervision via co-training~\cite{egomimic, qiu2025humanoidpolicyhuman}. However, these works have predominantly focused on table-top manipulation. To this end, we introduce \emph{Egocentric Mobile Manipulation} (\method{}), an end-to-end system to train mobile manipulation policies solely from \emph{robot static-manipulation} and \emph{human mobile-manipulation} data. While the robot data is captured via teleoperation, the human data is captured simply by wearing Project Aria glasses~\cite{engel2023projectarianewtool}, thus avoiding costly mobile manipulation teleoperation.
 
\method{} represents a step toward a new data paradigm in robot learning, where we imbue a robot with new skills by combining a set of robot teleoperation data with a more diverse pool of human demonstrations. In this work, our robot teleoperation data contains no mobile manipulation demonstrations, and we show that this skill can be effectively transferred from human data. \method{} presents a \emph{full-stack framework} consisting of 1) an action retargeting pipeline which translates human mobile manipulation actions to a bimanual robot with a differential-drive base, 2) a unified architecture designed for co-training on heterogeneous human and robot data, and 3) an auxiliary phase identification mechanism that modulates between navigation and manipulation modes during inference. Therefore, it can switch control modes between navigation and manipulation phases based on predictions, preventing unintended base drift during manipulation and out-of-distribution arm movements during navigation.
                                             
We evaluate \method{} across four real-world mobile manipulation tasks: \textit{table service}, \textit{handover wine}, \textit{grocery shopping} and \textit{push chair}. Our experiments reveal three key findings. First, \method{} achieves superior performance compared to baselines trained on teleoperated mobile robot data, showing that human demonstrations can replace costly mobile teleoperation. Second, we observe favorable scaling properties—each additional portion of human data yields greater performance than an equivalent portion of teleoperated robot data. Third, \method{} exhibits robust generalization, transferring navigation and manipulation skills to novel environments seen only in human demonstrations. These results open new avenues for scalable robotic learning in real-world environments.

\section{Related Work}
\label{sec:citations}

\textbf{Behavior Cloning (BC).} Behavior Cloning (BC) has emerged as an effective approach for robot learning, where policies are trained with direct supervised learning from expert demonstrations. Recent advances have shown remarkable results~\cite{rt22023arxiv, kim24openvla,black2410pi0,chi2024diffusionpolicy,zhao2023learning,o2024open}, including the promise of building general-purpose policies by learning from large-scale datasets~\cite{kim24openvla,o2024open,khazatsky2024droid}. In particular, research around large multi-embodiment datasets such as Open-X~\cite{o2024open} represents a significant milestone, demonstrating how models trained with diverse robot embodiments can acquire generalizable skills across tasks without task-specific engineering. However, most of these approaches focus on static manipulation tasks, with mobile manipulation remaining a significant challenge. 

\textbf{Learning for Mobile Manipulation.}

Building on successes in static manipulation, recent works have explored learning-based approaches for mobile manipulation that include skill primitives~\cite{sun2022fully, wu2023m, wu2023tidybot, kuang2025skillblender}, reinforcement learning with decomposed action spaces ~\cite{gu2022multi, yokoyama2023asc, xia2021relmogen, xiong2024adaptive}, and whole-body control objectives~\cite{fu2023deep, yang2024harmonic, hu2023causal, liu2024opt2skill, ben2025homie, wang2024hypermotion}. Unlike these approaches, end-to-end imitation learning enables mapping raw pixel information to whole-body actions, showing promising results through large-scale training~\cite{kim24openvla, ahn2022can, brohan2022rt, shafiullah2023bringing, pi0.5, moo2023arxiv}.
A critical challenge is to collect high-quality mobile manipulation demonstrations. Pioneering works~\cite{jiang2025behavior,fu2024humanplus} have developed new teleoperation systems to facilitate data collection, including tethered leader-follower system~\cite{fu2024humanplus}, VR headsets~\cite{chen2024arcapcollectinghighqualityhuman, lu2025mobiletelevisionpredictivemotionpriors}, motion-capture suits~\cite{cisneros2022team, dafarra2024icub3, wholebodygeomretarget}, smartphone-based control~\cite{pmlr-v164-wong22a}, kinesthetic teaching~\cite{MOMO-FORCE}, and full-body teleoperation for humanoid robots~\cite{fu2024humanplus}. However, despite these advances, collecting high-quality mobile manipulation data for diverse scenarios at scale remains a challenge. We propose to leverage egocentric human data (data collected with first-person perspective) captured by wearable devices as an alternative for scaling up imitation learning for mobile manipulation. 

\textbf{Robot Learning from Human Data.} 
Recent work has focused on two complementary themes—leveraging human videos to bootstrap robot learning and finetuning with reinforcement learning for robust policies. In manipulation, co-training on paired egocentric human and robot demonstrations has been shown to boost skill performance~\cite{egomimic}, while zero- or few-shot transfer from human videos can be enabled by image inpainting and motion-track priors~\cite{lepert2025phantomtrainingrobotsrobots, ren2025motion}. Building on these, hierarchical planners extract latent action sequences from humans and distill them via retargeters into whole-body controllers~\cite{wang2023mimicplay, fu2024humanplus, ImitationNet}. For navigation, policies become more reliable when imitation is fused with RL, either by converting keypoint matches into rewards or through behavior cloning bootstrapping~\cite{karnan2022voila, ramrakhya2023pirlnav}. Furthermore, inverse-dynamics models can pseudo-label passive egocentric video to distill intent-conditioned affordance subroutines~\cite{navsubroutine_egocentricvideos}. We aim to train mobile manipulation policies from human data in a unified learning framework.

\section{Hardware and Data Preliminaries}
\textbf{Egocentric Data Collection.} In this work, we leverage a wearable smart glass Meta Project Aria~\cite{engel2023projectarianewtool} as our main data collection platform. Echoing prior work~\cite{egomimic}, we believe Aria glasses are ideal for capturing egocentric human data due to their ergonomic design and machine perception capability provided by the Machine Perception Service (MPS). Specifically, we leverage Aria glasses to capture both \emph{exteroception} (wide-FOV egocentric RGB images) and \emph{proprioception} (hand tracking and global localization) data in human mobile manipulation behaviors. 

\begin{figure*}[t]
\centering
\includegraphics[width=0.95\linewidth]{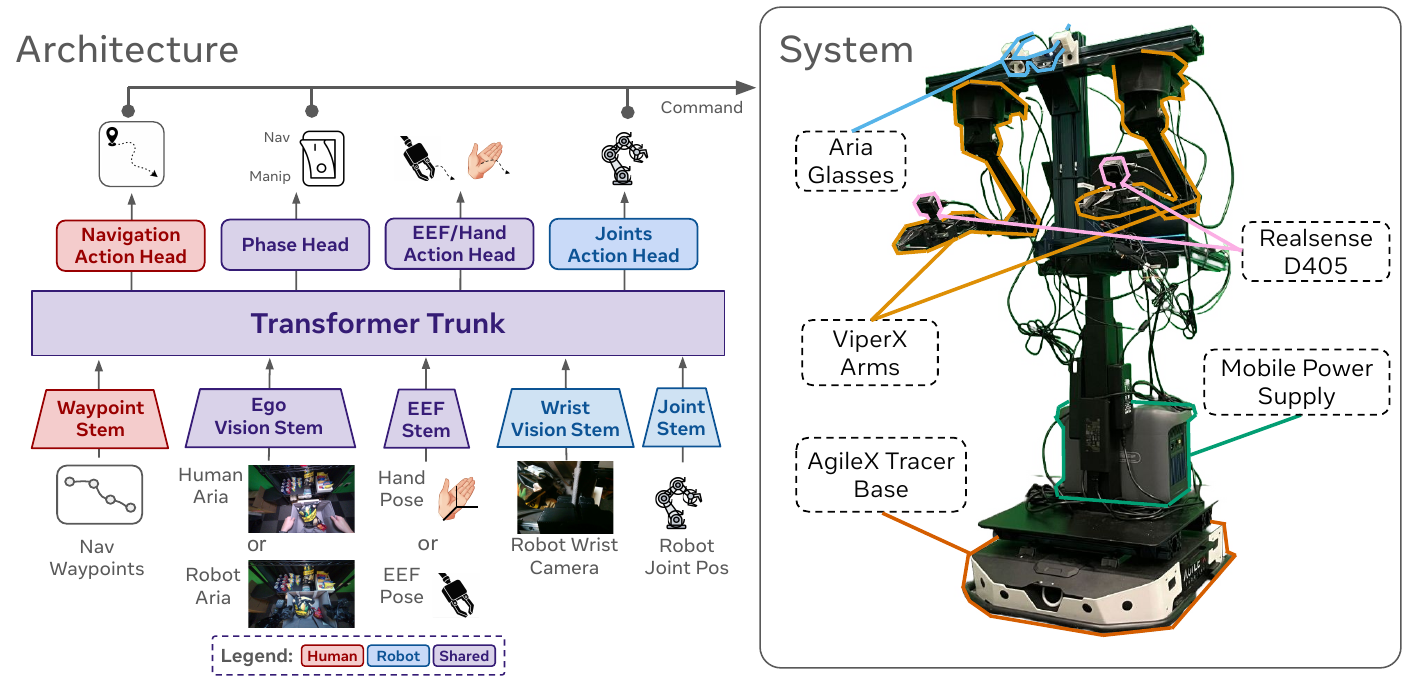}
\caption{Left: Architecture of joint human-robot policy learning framework, built on top of~\cite{wang2024hpt}. Our model processes heterogeneous human and robot data through \textit{stems} and decodes them through various action \textit{heads}. The navigation head is deployed on the robot during evaluation, demonstrating transfer without robot supervision. Right: Our custom low-cost bimanual mobile manipulator.}
\vspace{-1.5em}
\label{fig:method}
\end{figure*}

\textbf{Low-cost Bimanual Mobile Manipulator.} To effectively utilize egocentric human data for mobile manipulation, the robot hardware platform must resemble human sizes and kinematic workspaces. Drawing inspiration from the ``Eve'' robot platform introduced in EgoMimic \cite{egomimic}, we develop a low-cost custom mobile manipulator that comprises of two 6-DoF ViperX 300s mounted in an identical inverted configuration on a height-adjustable rig. The rig is mounted on an AgileX TRACER differential drive AGV platform, which is capable of moving up to 2m/s. The full system stands a maximum height of 1.75m. Similar to Eve, we propose to leverage Aria glasses as the main egocentric perception sensor for the robot and mount it in a way that emulates the hand-eye configuration of a human adult. This mitigates the human-robot camera device gap and reduces the sensor-manipulator kinematic gap. Each arm is equipped with an Intel Realsense D405 on its wrist to facilitate precise near-range manipulation. 

\textbf{Human and Robot Data Streams.} We collect distinct data streams from both human demonstrations ($\mathcal{D}_H$) and robot execution ($\mathcal{D}_R$). A key shared data stream is the egocentric RGB image $I_{\text{ego}}$, generated by the Aria glasses worn by the human or mounted on the robot. The robot provides RGB streams from its two wrist cameras, $I_{\text{wrist}}$. Proprioceptive data streams capture the state of each embodiment: For the human, we leverage the Aria Machine Perception Service (MPS)~\cite{engel2023aria} to estimate bimanual 3D hand poses $H_p \in SE(3) \times SE(3)$ and the 3D head pose $H_d \in SE(3)$ (in a SLAM-based world frame). For the robot, proprioception includes the state of its bimanual arms via joint positions $R_q \in \mathbb{R}^{2 \times 7}$ (including gripper state) and the corresponding end-effector poses $R_p \in SE(3) \times SE(3)$. These raw streams serve as the inputs to our system; their processing and transformation into policy inputs and action labels is a core algorithmic contribution detailed in Sec.~\ref{sec:retargeting}.

\section{EMMA: System and Algorithm}
\label{sec:methods}

\method{} is a scalable \emph{full-stack system}, which enables (1) \emph{direct transfer} of navigation skills from egocentric human data to a differential-drive mobile manipulator and (2) \emph{scaling up} full mobile manipulation policy performance by co-training on both human mobile manipulation data and robot static manipulation data (Fig.~\ref{fig:method}).

\subsection{Data Retargeting and Alignment}
\label{sec:retargeting}
A fundamental challenge in leveraging human data ($\mathcal{D}_H$) for robot learning is the embodiment gap, defined as the mismatch in physical morphology and movement kinematics between human and robot embodiments, which affects navigation and manipulation. Humans navigate omnidirectionally with decoupled head-gaze, whereas our robot uses a differential-drive base with kinematically constrained movements. Similarly, human hand motions ($H_p$) captured egocentrically differ kinematically and distributionally from robot end-effector motions ($R_p$). To enable effective knowledge transfer and co-training in an end-to-end imitation learning setting (Sec.~\ref{sec:cotraining}), we introduce two distinct data processing strategies: optimization-based retargeting for navigation and coordinate-space alignment for manipulation.

\textbf{Bridging Navigation Kinematic Gap.} We process the raw human head pose $H_d$ (Sec.~\ref{sec:methods}) to generate \emph{both} the ground-truth action labels for the policy and the contextual state input.

\begin{figure}[t]
\centering
\vspace{-1em}
\includegraphics[width=\linewidth]{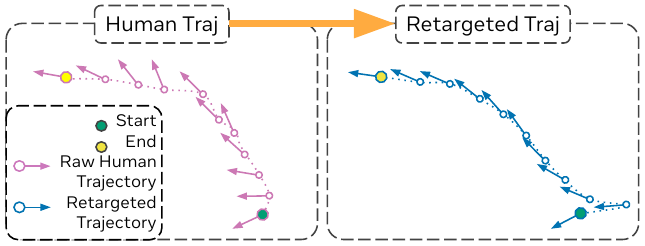}
\caption{Given the ground-plane projection of the human head trajectory, we optimize Eq.~\ref{eq:retargeting} to produce a smooth and executable path for our differential-drive robot that can be run directly and used as input for policy learning.}
\vspace{-1.5em}
\label{fig:retargeting}
\end{figure}

\emph{1. Navigation Action Labels:} The primary focus of our retargeting efforts is translating human navigation trajectories into ground-truth action labels suitable for our differential-drive robot (Fig.~\ref{fig:retargeting}). During human demonstrations, we extract 2D base poses $h_{\text{base}}^t = (x^t, y^t, \theta^t)$ by projecting the 3D head pose $H_d^t$ onto the ground plane. These poses form navigation waypoints that capture the human's intended path.
However, directly mapping this sequence of waypoints to robot base commands is ill-posed due to differential-drive constraints (the robot can only move in straight lines and circular arcs) and the fixed alignment between the robot's torso-mounted Aria sensor and its heading. To overcome this, we formulate an optimization problem: given a sequence of desired waypoints $\{h_{\text{base}}^k = (x_k^d, y_k^d, \theta_k^d)\}_{k=1}^K$ extracted from human trajectory, find velocity commands $\mathbf{z} = [(v_1, \omega_1), ..., (v_K, \omega_K)]$ that minimize:
\vspace{-0.5em}
\begin{align}
\min_{\mathbf{z}} \sum_{k=1}^{K} \bigg[ &\lambda_{\text{pos}} \|p_k(\mathbf{z}) - p_k^d\|_2^2 + \lambda_{\text{yaw}} \text{wrap}(\theta_k(\mathbf{z}) - \theta_k^d)^2 \nonumber \\
&+ \lambda_{\text{smooth}} \left((v_k - v_{k-1})^2 + (\omega_k - \omega_{k-1})^2\right) \bigg] \label{eq:retargeting}
\end{align}
\vspace{-0.5em}
subject to the differential-drive dynamics:
\begin{align}
x_{k+1} &= x_k + v_k \cos(\theta_k)\Delta t \\
y_{k+1} &= y_k + v_k \sin(\theta_k)\Delta t \\
\theta_{k+1} &= \theta_k + \omega_k\Delta t
\end{align}
\begin{equation}
\operatorname{wrap}(\alpha)=\big((\alpha+\pi)\bmod 2\pi\big)-\pi \in [-\pi,\pi).
\end{equation}
and constraints $-1.6 \leq v_k \leq 1.6$ m/s and $-1.5 \leq \omega_k \leq 1.5$ rad/s. Here, $p_k(\mathbf{z}) = (x_k, y_k)$ represents the robot position at step $k$ resulting from applying the velocity sequence $\mathbf{z}$, $p_k^d = (x_k^d, y_k^d)$ is the desired position from the human waypoint, and $\text{wrap}(\cdot)$ ensures angular differences are in $[-\pi, \pi]$. The weights $\lambda_{\text{pos}}=25$, $\lambda_{\text{yaw}}=2$, and $\lambda_{\text{smooth}}=1$ balance position tracking, heading alignment, and velocity smoothness respectively. This constrained optimization yields smooth, kinematically feasible trajectories that approximate human navigation patterns under differential-drive constraints.

\emph{2. Navigation Context:} To provide the policy with speed-invariant context for predicting navigation actions, we process the human head pose $H_d$ into a history of waypoints by projecting $H_d$ onto the ground plane to obtain $h_{\text{base}}^t = (x^t, y^t, \theta^t) \in SE(2)$. We then maintain a displacement-based waypoint history $\mathcal{W}_t = \{h_{\text{base}}^{t-k_i}\}_{i=1}^{K_h}$, where $K_h$ is the maximum number of historical waypoints. New waypoints are added when the displacement $\|h_{\text{base}}^t - h_{\text{base}}^{t-k_i}\|_2 \geq d_{\text{thresh}}$ (e.g., every 0.5m). This displacement-based sampling ensures consistent spatial resolution. For policy learning, these waypoints are transformed into the current egocentric frame as $\tilde{\mathcal{W}}_t = \{T_{ego}^{-1} \cdot h_{\text{base}}^{t-k_i}\}_{i=1}^{K_h}$, where $T_{ego}$ is the transformation from world to current egocentric coordinates. This egocentric waypoint history serves as a contextual input to the policy.

\textbf{Aligning Manipulation Action Data.} For manipulation, we address mismatches between human hand data ($H_p$) and robot end-effector data ($R_p$). Inspired by prior work like EgoMimic~\cite{egomimic}, we first unify coordinate frames by transforming all upper-body action chunks (both human and robot) into the reference frame of the camera \emph{at the time of observation}, using SLAM estimates for human data and hand-eye calibration for robot data. This makes predictions relative to the current view. Second, acknowledging persistent distributional gaps due to biomechanics and sensors, we apply Z-score normalization independently to the transformed pose and action data within each source ($\mathcal{D}_H$ and $\mathcal{D}_R$), using their respective dataset statistics.

\vspace{-1em}

\subsection{Human and Robot Data Co-training}
\label{sec:cotraining}

By retargeting human demonstrations into executable robot trajectories (Sec. IV-A), we bridge the embodiment gap, allowing the co-training loop to treat heterogeneous human ($D_H$) and robot ($D_R$) data as a single, semantically aligned stream for the shared Transformer trunk.

Inspired by recent works in cross-embodiment policy learning~\cite{wang2024hpt, liu2024rdt, ghosh2024octo}, we design an architecture based on a decoder-only Transformer with modality-specific input \emph{stems}, a shared \emph{trunk}, and multiple action and auxiliary output \emph{heads} (Fig.~\ref{fig:method}).

Our goal is to transfer semantics of navigation and manipulation phases from human data. In addition, we want both the manipulation action prediction and navigation action prediction to benefit from large-scale human data. This motivates an architecture that shares a majority of the learnable parameters between the human and robot embodiments to promote shared representation, inspired by EgoMimic \cite{egomimic}. 

\textbf{Stems.} Stems are shallow networks that encode raw observations from different modalities into a sequence of fixed-dimension tokens. Crucially, we employ a \emph{shared} vision stem for processing the main egocentric RGB images ($I_{ego}$) from the Aria glasses (human and robot) to enforce visual feature alignment. Separate stems handle inputs from the robot's wrist cameras ($I_{wrist}$), which does not exist in human data. 

\textbf{Trunk.} The Trunk is a standard multi-layer decoder-only transformer that processes the concatenated token sequences from all active stems. A sequence of $M$ learnable tokens, representing the action chunk length, is prepended to the input sequence constructed from all of the stems.

\textbf{Heads}. The heads are shallow MLPs that map the first $M$ output tokens from the trunk to the respective action spaces or auxiliary predictions. We define four heads for predicting robot bimanual joint actions ($ \mathbb{R}^{K \times 14}$), human cartesian end-effector actions {($\mathbb{R}^{K \times 3}$)}, robot base navigation actions ($(x, y, \omega) \in \mathbb{R}^{K \times 3}$), and, as an auxiliary output, the predicted task phase ($p \in \{0, 1\}^{K}$, see Sec.~\ref{sec:phase_labeling}).

\textbf{Co-training from heterogeneous sources.} The model is trained jointly on batches drawn from two sources: (1) the collected human mobile manipulation demonstrations $D_H$ (with navigation actions processed by the retargeting module described in Sec.~\ref{sec:retargeting}) and (2) static robot manipulation demonstrations $D_R$ (e.g., collected via teleoperation).
When processing a human data batch, the human proprioception, shared ego vision stem, human manipulation action head, navigation action head and the shared phase prediction head are active. For a robot batch, the robot proprioception, wrist image stem, shared ego vision stem, and robot manipulation action head are active. The shared trunk and ego vision stem are updated by all sources and modalities, forcing them to learn versatile representations. The navigation head primarily learns from the retargeted human data, transferring human navigation strategies to the robot. The complete architecture is illustrated in Fig.~\ref{fig:method}.

\subsection{Auxiliary Phase Identification and Control Modulation}
\label{sec:phase_labeling}
Mobile manipulation tasks naturally alternate between navigation and manipulation phases, requiring different control strategies. We introduce an unsupervised phase identification mechanism that automatically segments demonstrations and modulates control during deployment.

\textbf{Phase Detection.} We identify phases based on motion dynamics. For each frame, we compute the ratio of hand velocity to head velocity $r$ as detailed in Algorithm~\ref{alg:phase_id}. To assess the accuracy of this unsupervised method, we hand-annotate 10\% of uniformly sampled continuous segments from four tasks that require phase identification. We report \textbf{Mean over Frames (MoF)}, defined as the number of frames where the manual label matches the dataset label divided by the total number of sampled frames.

\begin{algorithm}[!ht]
\caption{Unsupervised Phase Identification}
\label{alg:phase_id}
\begin{algorithmic}[1]
\State \textbf{Input:} Head poses $\mathbf{p}_{head}$, hand poses $\mathbf{p}_{hand}$
\State \textbf{Output:} Phase labels $\phi \in \{0, 1\}^N$
\State $v_{head},v_{hand} \gets \|\Delta \mathbf{p}_{head}\| / \Delta t, \|\Delta \mathbf{p}_{hand}\| / \Delta t$
\State $r \gets v_{hand} / (v_{head} + \epsilon)$
\State $\mathcal{M} \gets \{i : r_i > \tau_{ratio} \land v_{head,i} < \tau_{head}\}$
\State Fit GMM($K$) on $\{\mathbf{p}_{head,i} : i \in \mathcal{M}\}$
\State $\phi_i \gets \text{GMM.pdf}(\mathbf{p}_{head,i}) < \tau_{pdf}$
\end{algorithmic}
\end{algorithm}

We empirically selected $\tau_{ratio}=2.0$, $\tau_{head}=0.4m/s$, and $\tau_{duration}=30$ based on the Table Service task. Consistently high MoF scores across tasks $(all \ge 0.92)$ confirm the system's robustness to these specific threshold choices. To spatially localize $K$ manipulation zones, we fit a Gaussian Mixture Model (GMM) with $K=2$ components to head positions during these high-ratio periods. Each frame is classified as manipulation (phase 0) if its probability density under the GMM exceeds $\tau_{pdf}$, otherwise it is labeled as navigation (phase 1). GMM provides direct probability density estimates needed for classification without requiring labeled training data, while being computationally efficient and interpretable for spatial clustering.

\begin{table}[!ht]
\centering
\begin{tabular}{|lcccc|}
\toprule
\textbf{Task} & \shortstack{Table \\ Service} & \shortstack{Handover \\ Wine} & \shortstack{Handover Wine \\ Scene Gen} & \shortstack{Grocery \\ Shopping} \\
\midrule
MoF & 0.962 & 0.965 & 0.929 & 0.963 \\
\bottomrule
\end{tabular}
\end{table}

\textbf{Phase-Aware Control:} During deployment, predicted phases modulate the navigation action chunk to prevent unintended movements:
\begin{itemize}
\item \textbf{Manipulation phases:} Navigation actions interpolate from zero to the first future navigation waypoint.
\item \textbf{Navigation phases:} Any future manipulation-phase positions are replaced with the current navigation endpoint.
\end{itemize}
This phase-aware modulation removes the base action noise introduced from head movements during manipulation and ensures a smooth switch between the two phases.

\section{Experiments}
\label{sec:expriments}
\begin{figure*}[htbp]
    \centering
    \includegraphics[width=\textwidth]{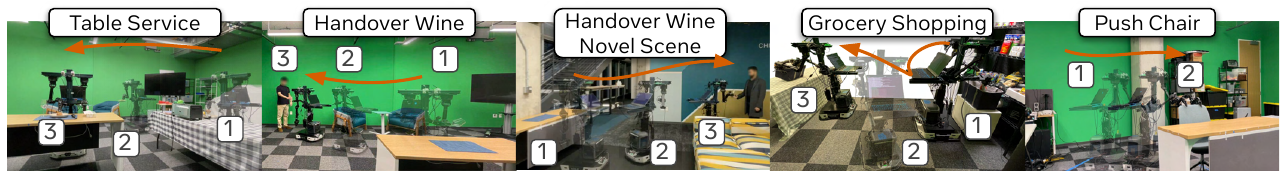}
    \caption{\textbf{Time-lapse} of all five tasks (variants) showing the task flow. \method{} is capable of learning tasks from human-robot interaction to precise bimanual  coordination. Numbers indicate chronological sequence.}
    \label{fig:main_qualitative}
    \vspace{-1em}
\end{figure*}

We aim to validate three key hypotheses. \textbf{H1:} EMMA can achieve performance comparable to systems trained on teleoperated mobile manipulation data. \textbf{H2:} Key design decisions of EMMA improve downstream task performance and robustness. \textbf{H3:} Given an initial amount of static robot manipulation data, it is more valuable to collect additional human mobile manipulation over mobile robot teleoperation data. We evaluate these hypotheses through four long horizon mobile manipulation tasks (Fig.~\ref{fig:main_qualitative}).

\begin{figure*}[htbp]
    \centering
    \includegraphics[width=\textwidth]{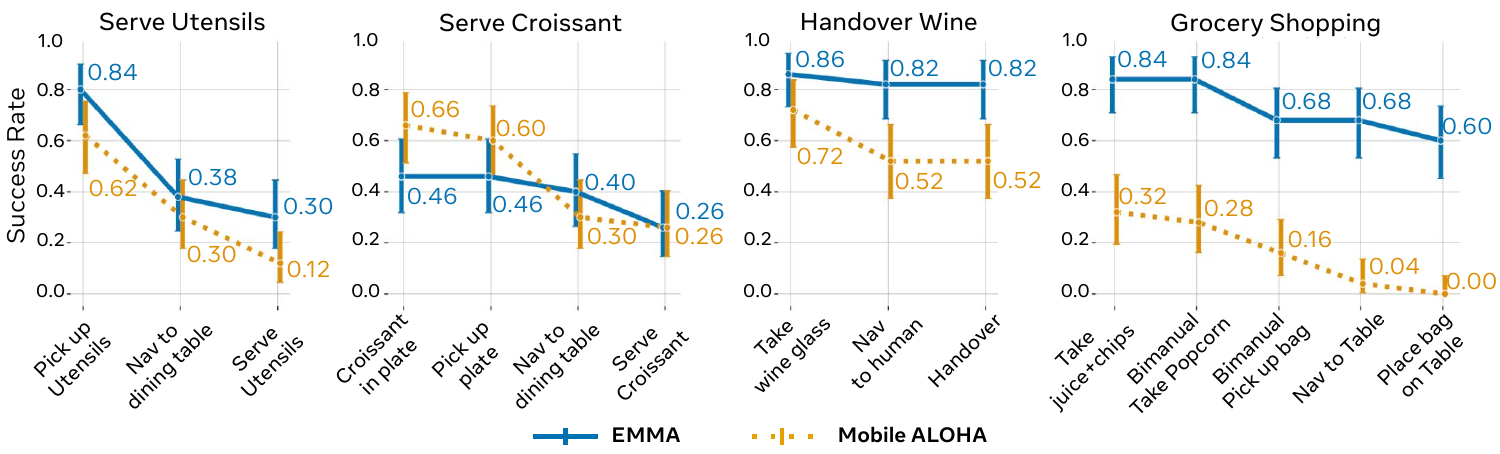}
    \caption{Cumulative success rates across subtasks for three mobile manipulation tasks. EMMA (blue), trained without mobile teleoperation data, significantly outperforms \baseline{} (orange) on \textit{Grocery Shopping} and \textit{Handover Wine} tasks (p $<$ 0.05). \textit{Table Service} variants show comparable performance. Error bars represent 95\% Clopper–Pearson confidence intervals with N = 50 trials.}
    \label{fig:success-rate}
    \vspace{-2em}
\end{figure*}

\textit{Table Service.} Two tables are set 2m apart. The kitchen table has an oven with four croissants, a plate, and wrapped utensils. The dining table has a mat with a wine glass in the corner. The robot picks up the utensils and navigates to the dining table, placing them on the left side. It then returns to the kitchen table, picks and places a croissant onto the plate, and navigates back to place the plate in the center of the dining mat, avoiding the wine glass.

\textit{Handover Wine.} The robot picks up a wine glass randomly placed on a tabletop dining mat. It turns right and navigates toward a human standing in a 3m×3m area. At a safe range, the robot hands the wine glass to the human's right hand, testing precise navigation and state coverage transfer from human data.

\textit{Grocery Shopping.} The robot faces a grocery shelf and simultaneously grabs a juice pouch from the left and a chip bag from the right, placing them into a shopping bag. It then uses both arms to pick a large popcorn bag from the center shelf and add it to the bag. Finally, the robot lifts the shopping bag and navigates to a table behind it, testing long-horizon bimanual manipulation and navigation.

\textit{Push Chair.} Grasp the backrest of an office chair and push it against a table. This task requires coupled arm-based motion to move and align the orientation of the chair. Unlike sequential tasks, \method{} here operates \emph{without phase-switching} to demonstrate capacity for unified full-body coordination. We co-train a base \method{} model using 10 min of mobile robot data and 20 min of human demonstrations to: (1) compare against \baseline{} trained on 30 min of robot-only data and (2) compare our unified policy against an ablation of separate navigation/manipulation models.

\textbf{Baselines.} We implement \textbf{\baseline{}}~\cite{fu2024mobile} as our primary baseline to compare against teleoperated mobile robot data. Critically, it exemplifies the exact data collection paradigm—expensive mobile teleoperation—that \method{} aims to replace with human egocentric data. To ensure fair comparison, we modify it to the same HPT~\cite{wang2024hpt} backbone used in EMMA, isolating the comparison to data sources (human vs. teleoperation) rather than architectural differences. This baseline receives identical input modalities (egocentric RGB, wrist cameras, proprioception) and outputs robot joint actions $R_q \in \mathbb{R}^{14}$ plus base velocities $(v, \omega) \in \mathbb{R}^2$ recorded from the AgileX Tracer wheel encoders.

In addition, we evaluate two ablation baselines. \textbf{EMMA w/o action retargeting} replaces our optimization-based retargeting (Eq.~\ref{eq:retargeting}) with raw human navigation actions, testing whether kinematic alignment is necessary for successful transfer. \textbf{EMMA w/o phase identification} removes the phase identification mechanism that modulates control during deployment, allowing both arm and base to move simultaneously throughout execution.

\textbf{Evaluation protocols.} We conduct 50 trials per task/model variant, recording error bars with 95\% Clopper-Pearson confidence interval. To compare the success rates between EMMA and our baseline, we use a two-proportion Z-test. We claim statistically significant outperformance for a model if p $<$ 0.05. Each trial is limited to 2 minutes. Any collisions are recorded as a failure for the corresponding subtask phase. All experiments use identical environmental conditions and object positions. We conducted \emph{1150 mobile manipulation rollout} evaluations in total.

\begin{figure}[t!]
    \centering
    \includegraphics[width=\linewidth]{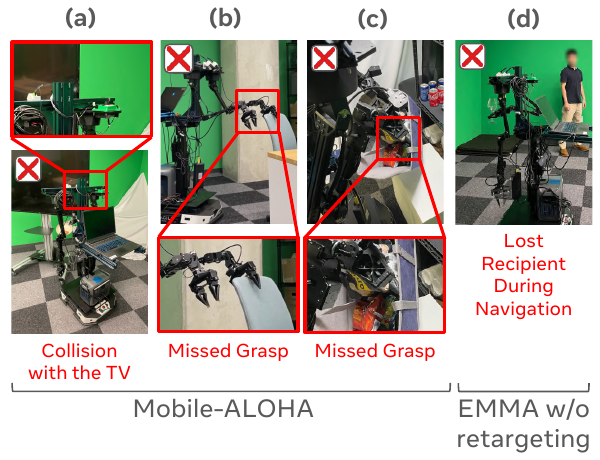}
    \caption{Representative failure cases of \baseline{} (a-c) and \method{} w/o retargeting (d). Analysis in Sec.~\ref{ssec:result}.}
    \label{fig:failure-cases}
    \vspace{-1.5em}
\end{figure}

\begin{figure}[!ht]
    \centering
    \includegraphics[width=\linewidth]{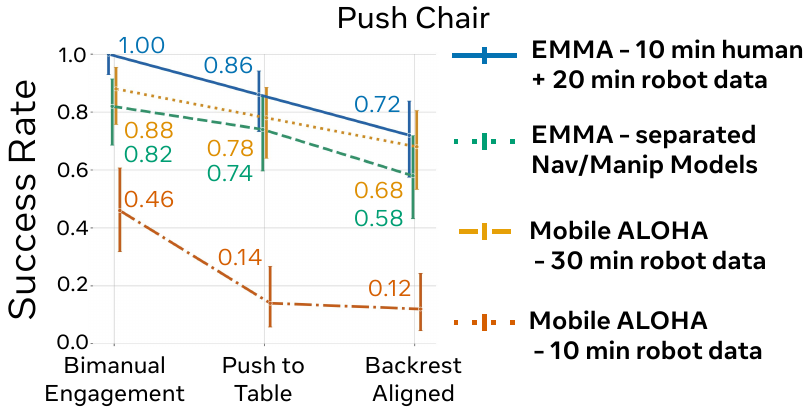}
    \caption{\textbf{Coupled arm-base motion task} (\textit{Push Chair}). Our unified \method{} (10m robot + 20m human) scales comparably to \baseline{} (30m robot) and outperforms separate navigation/manipulation models, while excelling at the \textbf{initial bimanual engagement}.}
    \vspace{-1em}
    \label{fig:push-chair}
\end{figure}

\subsection{Main Results}
\label{ssec:result}

\textbf{EMMA achieves favorable performance compared to systems trained on teleoperated mobile robot data (H1).}
EMMA consistently outperforms teleoperation-based baselines across a range of mobile-manipulation tasks by leveraging egocentric human demonstrations (Fig.~\ref{fig:success-rate}). This quantitative gap is reinforced by qualitative evidence, as the \baseline{} model exhibits frequent critical failures (Fig.~\ref{fig:failure-cases}): These include (a) navigation collisions, such as hitting the TV, and (b) flawed bimanual grasps, which result in incorrectly pushing the chair. In the \textit{Handover Wine} task, with the same 50 demos of static robot data, replacing one hour of teleoperated mobile robot data with one hour of human mobile manipulation yields an 82\% success rate—a 30\% increase over teleoperated robot data, which had a 52\% success rate. This improvement demonstrates the benefit of learning from egocentric human-human interaction. Similarly, for the \textit{Push Chair} task that requires coupled arm-base movements, adding 20 min of human data to 10 min of mobile robot data allows \method{} to match the performance of \baseline{} trained on 30 min of pure robot teleoperation, (Fig.~\ref{fig:push-chair}). Leveraging egocentric human mobile manipulation data has also shown improvement in bimanual tasks found in the \textit{Grocery Shopping} task. Lastly, using egocentric human mobile manipulation data has shown comparable performance to teleoperated mobile robot data on long-horizon \textit{Table Service} task, a task that requires long-distance point-to-point navigation, where robot-teleoperated methods benefit from zero perspective gap. Taken together, these results show that egocentric human mobile manipulation data matches or even improves the value of equivalent robot teleoperation data and delivers safer, more reliable, and more scalable policies across diverse mobile-manipulation challenges.

\begin{figure}[t!]
    \centering
    \includegraphics[width=\linewidth]{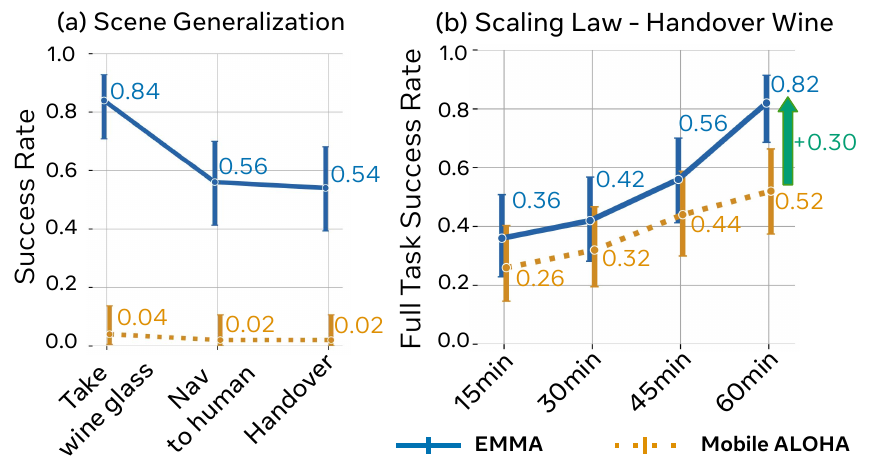}
    \caption{(a) \method{} generalizes to an unseen scene with 54\% full task success rate. (b) For \textit{Handover Wine} task, starting with a fixed amount of static manipulation data, we show that adding more human full-body motion data for \method{} (blue) yields greater performance gains compared to adding mobile robot teleoperation data collected in an equivalent amount of time for \baseline{} (orange). The performance gap expands from 10\% to 30\% as data increases from 15 to 60 minutes.}
    \label{fig:scaling-law-and-scene-gen}
    \vspace{-1em}
\end{figure}

\textbf{EMMA generalizes to unseen scenes.} We evaluate scene generalization by testing EMMA in a novel environment seen only through human demonstrations. Specifically, we collect 30 minutes of human demonstrations in a new spatial layout where the wine recipient stands randomly within a larger 5m $\times$ 2m area. Despite never seeing robot data from this environment, EMMA achieves 54\% success rate (Fig.~\ref{fig:scaling-law-and-scene-gen} (a)), demonstrating two key capabilities: (1) navigation behaviors learned from human data naturally adapt to the expanded spatial configuration, and (2) manipulation skills remain robust to visual domain shifts from the new environment. In contrast, Mobile ALOHA trained exclusively on teleoperated data from the original environment is unable to complete the initial grasp due to the environmental changes (Fig.~\ref{fig:scaling-law-and-scene-gen} (a)). This experiment validates that human data provides superior generalization through its inherent diversity compared to lab-constrained teleoperation.

\begin{figure}[htbp]
    \centering
    \includegraphics[width=\linewidth]{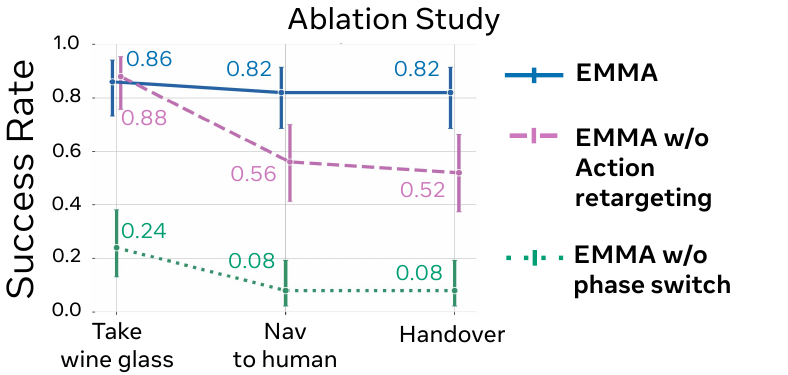}
    \caption{\textbf{Ablation study} on the \textit{Handover Wine} task. Removing either retargeting or phase switch causes significant performance drop.}
    \label{fig:ablation}
    \vspace{-1em}
\end{figure}

\textbf{Ablation Studies (H2).}
Our ablation experiments on the \textit{Handover Wine} task (Fig.~\ref{fig:ablation}) reveal the critical importance of both key components. Without kinematic retargeting, the success rate drops by 30\%, as the robot frequently loses track of the human recipient when executing kinematically infeasible trajectories. The phase identification mechanism proves equally crucial—its removal causes complete task failure due to unintended base movements during grasping and unnecessary arm motions during navigation, resulting in collisions and dropped objects. These results validate that both human-robot embodiment alignment and learned phase modulation are essential for successful human-to-robot skill transfer in mobile manipulation. On the \textit{Push Chair} task, a unified \method{} policy (10 min robot + 20 min human) outperforms separate navigation and manipulation models on precise bimanual chair grasp and continuous engagement.  (Fig.~\ref{fig:push-chair}), indicating that single-policy co-training better exploits human data for arm-base coupling. These results validate that our design choices are essential for successful human-to-robot skill transfer. Our observed representative failure modes further reinforce this (Fig.~\ref{fig:failure-cases}): (c) without phase switching, \baseline{} makes unintended base motions that hinder precise manipulation. (d) Without kinematic retargeting, executing raw human waypoints yields infeasible trajectories.

\textbf{Human data scales mobile manipulation performance more efficiently compared to teleoperated robot data (H3)}. Keeping one hour of static robot manipulation data fixed, EMMA’s success rate climbs steadily from 0.36 to 0.82 as we increase human mobile manipulation data from 15 minutes to one hour (Fig.~\ref{fig:scaling-law-and-scene-gen} (b)). \baseline{} also improves from 0.26 to 0.52 when its robot teleoperation data grows from 15 minutes to one hour, but remains consistently below \method{}. This demonstrates that additional human mobile manipulation demonstrations yield substantially greater returns than equivalent increases in mobile robot teleoperation data. 
\section{Limitations}

Despite the strengths of \method{}, our approach inherits several inherent limitations from learning mobile manipulation primarily through human demonstrations.
First, the framework assumes that the visual and spatial distributions encountered during robot deployment lie within, or close to, those seen in human demonstrations. This assumption may break down when the kinematic or viewpoint differences between human and robot embodiments become substantial.
Second, the direct transfer of human navigation strategies is insufficient for tasks requiring tightly coupled arm–base coordination.  Nevertheless, we showed that human data provided significant scaling when combined with even limited mobile robot data.

\section{Conclusion}
\label{sec:conclusion}

In conclusion, we presented \method{}, a novel framework that enables mobile manipulation without requiring expensive mobile teleoperation. We find that \method{} outperforms \baseline{} style teleoperation with equivalent data collection time and demonstrates superior scaling properties. Further, we ablate our key design decisions and observe that both the motion retargeter and the phase identification module are integral to downstream performance. Overall, we have demonstrated the possibility of scaling robot mobile manipulation performance with egocentric human data. We hope this result spurs new research and informs future researchers of potential opportunities and challenges. 

\section{Acknowledgment}

This work was partially supported by the Samsung Research America LEAP-U Program, the Industrial Technology Innovation Program (P0028404) of the Ministry of Industry, Trade and Energy of Korea, and a research gift from Meta Platforms, Inc.

\bibliographystyle{IEEEtran}
\bibliography{bibliography/references}

@inproceedings{fu2024mobile,
  author    = {Fu, Zipeng and Zhao, Tony Z. and Finn, Chelsea},
  title     = {Mobile ALOHA: Learning Bimanual Mobile Manipulation with Low-Cost Whole-Body Teleoperation},
  booktitle = {Proc. Conf. Robot Learn. (CoRL)},
  year      = {2024}
}

@article{qiu2025humanoidpolicyhuman,
title={Humanoid Policy \~{} Human Policy},
author={Qiu, Ri‑Zhao and others},
journal={arXiv preprint arXiv:2503.13441},
year={2025}
}

@misc{bharadhwaj2024track2actpredictingpointtracks,
  author = {Bharadhwaj, Homanga and others},
  title  = {Track2Act: Predicting Point Tracks from Internet Videos enables Generalizable Robot Manipulation},
  year   = {2024}
}

@misc{xu2024flowcrossdomainmanipulationinterface,
  author = {Xu, Mengda and others},
  title  = {Flow as the Cross-Domain Manipulation Interface},
  year   = {2024},
  note   = {arXiv:2407.15208}
}

@misc{wen2024anypointtrajectorymodelingpolicy,
  author = {Wen, Chuan and others},
  title  = {Any-point Trajectory Modeling for Policy Learning},
  year   = {2024},
  note   = {arXiv:2401.00025}
}

@misc{engel2023projectarianewtool,
  author = {Engel, Jakob and others},
  title  = {Project Aria: A New Tool for Egocentric Multi-Modal {AI} Research},
  year   = {2023},
  note   = {arXiv:2308.13561}
}

@misc{nair2022r3muniversalvisualrepresentation,
  author = {Nair, Suraj and others},
  title  = {{R3M}: A Universal Visual Representation for Robot Manipulation},
  year   = {2022},
  note   = {arXiv:2203.12601}
}

@article{zhao2023learning,
  author  = {Zhao, Tony Z. and others},
  title   = {Learning fine-grained bimanual manipulation with low-cost hardware},
  journal = {arXiv preprint arXiv:2304.13705},
  year    = {2023}
}

@article{khazatsky2024droid,
  author  = {Khazatsky, Alexander and others},
  title   = {Droid: A large-scale in-the-wild robot manipulation dataset},
  journal = {arXiv preprint arXiv:2403.12945},
  year    = {2024}
}

@inproceedings{o2024open,
  author    = {O'Neill, Abby and others},
  title     = {Open X-Embodiment: Robotic Learning Datasets and {RT-X} Models},
  booktitle = {Proc. IEEE Int. Conf. Robot. Autom. (ICRA)},
  pages     = {6892--6903},
  year      = {2024}
}

@inproceedings{rt22023arxiv,
  author    = {Brohan, Anthony and others},
  title     = {{RT-2}: Vision-Language-Action Models Transfer Web Knowledge to Robotic Control},
  booktitle = {arXiv preprint arXiv:2307.15818},
  year      = {2023}
}

@inproceedings{wang2024hpt,
  author    = {Wang, Lirui and others},
  title     = {Scaling Proprioceptive-Visual Learning with Heterogeneous Pre-trained Transformers},
  booktitle = {Proc. NeurIPS},
  year      = {2024}
}

@article{jiang2025behavior,
  author  = {Jiang, Yunfan and others},
  title   = {{BEHAVIOR} Robot Suite: Streamlining Real-World Whole-Body Manipulation for Everyday Household Activities},
  journal = {arXiv preprint arXiv:2503.05652},
  year    = {2025}
}

@article{wang2023mimicplay,
  author  = {Wang, Chen and others},
  title   = {MimicPlay: Long-horizon imitation learning by watching human play},
  journal = {arXiv preprint arXiv:2302.12422},
  year    = {2023}
}

@inproceedings{karnan2022voila,
  author    = {Karnan, Haresh and others},
  title     = {Voila: Visual-observation-only imitation learning for autonomous navigation},
  booktitle = {Proc. IEEE Int. Conf. Robot. Autom. (ICRA)},
  pages     = {2497--2503},
  year      = {2022}
}

@article{ren2025motion,
  author  = {Ren, Juntao and others},
  title   = {Motion Tracks: A Unified Representation for Human-Robot Transfer in Few-Shot Imitation Learning},
  journal = {arXiv preprint arXiv:2501.06994},
  year    = {2025}
}

@misc{lepert2025phantomtrainingrobotsrobots,
  author = {Lepert, Marion and others},
  title  = {Phantom: Training Robots Without Robots Using Only Human Videos},
  year   = {2025},
  note   = {arXiv:2503.00779}
}

@article{black2410pi0,
  author  = {Black, Kevin and others},
  title   = {$\pi$0: A Vision-Language-Action Flow Model for General Robot Control},
  year    = {2024},
  journal = {arXiv:2410.24164}
}

@article{chi2024diffusionpolicy,
  author  = {Chi, Cheng and others},
  title   = {Diffusion Policy: Visuomotor Policy Learning via Action Diffusion},
  journal = {Int. J. Robot. Res.},
  year    = {2024}
}

@misc{egomimic,
  author = {Kareer, Simar and others},
  title  = {EgoMimic: Scaling Imitation Learning via Egocentric Video},
  year   = {2024},
  note   = {arXiv:2410.24221}
}

@article{kim24openvla,
  author  = {Kim, Moo Jin and others},
  title   = {{OpenVLA}: An Open-Source Vision-Language-Action Model},
  journal = {arXiv preprint arXiv:2406.09246},
  year    = {2024}
}

@misc{pi0.5,
  author = {{Physical Intelligence} and others},
  title  = {$\pi_{0.5}$: A Vision-Language-Action Model with Open-World Generalization},
  year   = {2025},
  note   = {arXiv:2504.16054}
}

@inproceedings{fu2024humanplus,
  author    = {Fu, Zipeng and Zhao, Qingqing and Wu, Qi and Wetzstein, Gordon and Finn, Chelsea},
  title     = {HumanPlus: Humanoid Shadowing and Imitation from Humans},
  booktitle = {Proc. Conf. Robot Learn. (CoRL)},
  year      = {2024}
}

@inproceedings{ramrakhya2023pirlnav,
  author    = {Ramrakhya, Ram and others},
  title     = {{PIRLNav}: Pretraining with Imitation and {RL} Finetuning for {ObjectNav}},
  booktitle = {Proc. CVPR},
  year      = {2023}
}

@misc{navsubroutine_egocentricvideos,
  author = {Kumar, Ashish and others},
  title  = {Learning Navigation Subroutines from Egocentric Videos},
  year   = {2019},
  note   = {arXiv:1905.12612}
}

@inproceedings{ImitationNet,
  author    = {Yan, Yashuai and others},
  booktitle = {Proc. IEEE-RAS Int. Conf. Humanoid Robots},
  title     = {ImitationNet: Unsupervised Human-to-Robot Motion Retargeting via Shared Latent Space},
  year      = {2023},
  pages     = {1--8}
}

@inproceedings{MOMO-FORCE,
  author    = {Yang, Taozheng and others},
  booktitle = {Proc. IEEE/RSJ Int. Conf. Intell. Robots Syst. (IROS)},
  title     = {{MOMA-Force}: Visual-Force Imitation for Real-World Mobile Manipulation},
  year      = {2023},
  pages     = {6847--6852}
}

@inproceedings{pmlr-v164-wong22a,
  author    = {Wong, Josiah and others},
  title     = {Error-Aware Imitation Learning from Teleoperation Data for Mobile Manipulation},
  booktitle = {Proc. 5th Conf. Robot Learn. (CoRL)},
  pages     = {1367--1378},
  year      = {2022}
}

@inproceedings{cisneros2022team,
  author    = {Cisneros, R. and others},
  title     = {Team {JANUS} humanoid avatar: A cybernetic avatar to embody human telepresence},
  booktitle = {Proc. RSS Workshop},
  volume    = {3},
  year      = {2022}
}

@article{dafarra2024icub3,
  author  = {Dafarra, Stefano and others},
  title   = {{iCub3} avatar system: Enabling remote fully immersive embodiment of humanoid robots},
  journal = {Sci. Robot.},
  volume  = {9},
  number  = {86},
  year    = {2024}
}

@inproceedings{wholebodygeomretarget,
  author    = {Darvish, Kourosh and others},
  booktitle = {Proc. IEEE-RAS Int. Conf. Humanoid Robots},
  title     = {Whole-Body Geometric Retargeting for Humanoid Robots},
  year      = {2019},
  pages     = {679--686}
}

@misc{chen2024arcapcollectinghighqualityhuman,
  author = {Chen, Sirui and others},
  title  = {{ARCap}: Collecting High-quality Human Demonstrations for Robot Learning with Augmented Reality Feedback},
  year   = {2024},
  note   = {arXiv:2410.08464}
}

@misc{lu2025mobiletelevisionpredictivemotionpriors,
  author = {Lu, Chenhao and others},
  title  = {Mobile-TeleVision: Predictive Motion Priors for Humanoid Whole-Body Control},
  year   = {2025},
  note   = {arXiv:2412.07773}
}

@article{ahn2022can,
  author  = {Ahn, Michael and others},
  title   = {Do as I Can, Not as I Say: Grounding Language in Robotic Affordances},
  journal = {arXiv preprint arXiv:2204.01691},
  year    = {2022}
}

@article{brohan2022rt,
  author  = {Brohan, Anthony and others},
  title   = {{RT-1}: Robotics Transformer for Real-World Control at Scale},
  journal = {arXiv preprint arXiv:2212.06817},
  year    = {2022}
}

@article{shafiullah2023bringing,
  author  = {Shafiullah, Nur Muhammad Mahi and others},
  title   = {On Bringing Robots Home},
  journal = {arXiv preprint arXiv:2311.16098},
  year    = {2023}
}

@inproceedings{sun2022fully,
  author    = {Sun, Charles and others},
  title     = {Fully autonomous real-world reinforcement learning with applications to mobile manipulation},
  booktitle = {Proc. Conf. Robot Learn. (CoRL)},
  pages     = {308--319},
  year      = {2022}
}

@inproceedings{wu2023m,
  author    = {Wu, Bohan and Martin-Martin, Roberto and Fei-Fei, Li},
  title     = {M-ember: Tackling long-horizon mobile manipulation via factorized domain transfer},
  booktitle = {Proc. IEEE Int. Conf. Robot. Autom. (ICRA)},
  pages     = {11690--11697},
  year      = {2023}
}

@article{wu2023tidybot,
  author  = {Wu, Jimmy and others},
  title   = {Tidybot: Personalized robot assistance with large language models},
  journal = {Auton. Robots},
  volume  = {47},
  number  = {8},
  pages   = {1087--1102},
  year    = {2023}
}

@article{gu2022multi,
  author  = {Gu, Jiayuan and others},
  title   = {Multi-skill mobile manipulation for object rearrangement},
  journal = {arXiv preprint arXiv:2209.02778},
  year    = {2022}
}

@article{yokoyama2023asc,
  author  = {Yokoyama, Naoki and others},
  title   = {{ASC}: Adaptive skill coordination for robotic mobile manipulation},
  journal = {IEEE Robot. Autom. Lett.},
  volume  = {9},
  number  = {1},
  pages   = {779--786},
  year    = {2023}
}

@inproceedings{xia2021relmogen,
  author    = {Xia, Fei and others},
  title     = {RelMoGen: Integrating motion generation in reinforcement learning for mobile manipulation},
  booktitle = {Proc. IEEE Int. Conf. Robot. Autom. (ICRA)},
  pages     = {4583--4590},
  year      = {2021}
}

@inproceedings{fu2023deep,
  author    = {Fu, Zipeng and Cheng, Xuxin and Pathak, Deepak},
  title     = {Deep whole-body control: learning a unified policy for manipulation and locomotion},
  booktitle = {Proc. Conf. Robot Learn. (CoRL)},
  pages     = {138--149},
  year      = {2023}
}

@inproceedings{yang2024harmonic,
  author    = {Yang, Ruihan and others},
  title     = {Harmonic mobile manipulation},
  booktitle = {Proc. IEEE/RSJ Int. Conf. Intell. Robots Syst. (IROS)},
  pages     = {3658--3665},
  year      = {2024}
}

@article{hu2023causal,
  author  = {Hu, Jiaheng and Stone, Peter and Mart\'in-Mart\'in, Roberto},
  title   = {Causal policy gradient for whole-body mobile manipulation},
  journal = {arXiv preprint arXiv:2305.04866},
  year    = {2023}
}

@article{liu2024rdt,
  author  = {Liu, Songming and others},
  title   = {{RDT-1B}: A Diffusion Foundation Model for Bimanual Manipulation},
  journal = {arXiv preprint arXiv:2410.07864},
  year    = {2024}
}

@inproceedings{ghosh2024octo,
  author    = {{Octo Model Team} and others},
  title     = {Octo: An Open-Source Generalist Robot Policy},
  booktitle = {Proc. Robot.: Sci. Syst. (RSS)},
  year      = {2024}
}

@article{engel2023aria,
  author  = {Engel, J. and others},
  title   = {Project Aria: A New Tool for Egocentric Multi-Modal {AI} Research},
  journal = {arXiv preprint arXiv:2308.13561},
  year    = {2023}
}

@article{xiong2024adaptive,
  author  = {Xiong, Haoyu and Mendonca, Russell and Shaw, Kenneth and Pathak, Deepak},
  title   = {Adaptive Mobile Manipulation for Articulated Objects In the Open World},
  journal = {arXiv preprint arXiv:2401.14403},
  year    = {2024}
}

@article{liu2024opt2skill,
  author  = {Liu, Fukang and others},
  title   = {Opt2Skill: Imitating Dynamically-feasible Whole-Body Trajectories for Versatile Humanoid Loco-Manipulation},
  journal = {IEEE Robot. Autom. Lett.},
  year    = {2025}
}

@inproceedings{moo2023arxiv,
  author    = {Austin Stone and others},
  title     = {Open-World Object Manipulation using Pre-Trained Vision-Language Models},
  booktitle = {arXiv preprint},
  year      = {2023}
}

@article{ben2025homie,
  author  = {Ben, Qingwei and others},
  title   = {HOMIE: Humanoid Loco-Manipulation with Isomorphic Exoskeleton Cockpit},
  journal = {arXiv preprint arXiv:2502.13013},
  year    = {2025}
}

@inproceedings{wang2024hypermotion,
  author    = {Wang, Jin and others},
  title     = {HYPERmotion: Learning Hybrid Behavior Planning for Autonomous Loco-manipulation},
  booktitle = {Proc. 8th Annu. Conf. Robot Learn. (CoRL)},
  year      = {2024}
}

@article{kuang2025skillblender,
  author  = {Kuang, Yuxuan and others},
  title   = {SkillBlender: Towards Versatile Humanoid Whole-Body Loco-Manipulation via Skill Blending},
  journal = {arXiv preprint arXiv:2506.09366},
  year    = {2025}
}

\end{document}